\documentclass[11pt, oneside]{article}   	
\usepackage{geometry}                		
\usepackage{graphicx}				
\usepackage{amssymb}
\usepackage{amstext}
\usepackage{amsmath}
\usepackage{multirow}
\usepackage{url}

\title{Extremely Low Bit Neural Network: Squeeze the Last Bit Out with ADMM}
\author{Cong Leng, Zesheng Dou, Hao Li, Shenghuo Zhu, Rong Jin\\
Alibaba Group, Hang Zhou, China \\
\{lengcong.lc, zesheng.dzs, lihao.lh, shenghuo.zhu, jinrong.jr\}@alibaba-inc.com}
\date{}							

\begin{document}
\maketitle
\begin{abstract}
Although deep learning models are highly effective for various learning tasks, their high computational costs prohibit the deployment to scenarios where either memory or computational resources are limited. In this paper, we focus on compressing and accelerating deep models with network weights represented by very small numbers of bits, referred to as {\bf extremely low bit neural network}. We model this problem as a discretely constrained optimization problem. Borrowing the idea from Alternating Direction Method of Multipliers (ADMM), we decouple the continuous parameters from the discrete constraints of network, and cast the original hard problem into several subproblems. We propose to solve these subproblems using extragradient and iterative quantization algorithms that lead to considerably faster convergency compared to conventional optimization methods. Extensive experiments on image recognition and object detection verify that the proposed algorithm is more effective than state-of-the-art approaches when coming to extremely low bit neural network.
\end{abstract}

\section{Introduction}
These years have witnessed the success of convolutional neural networks (CNNs) in a wide range computer vision tasks, such as image classification, object detection and segmentation. The success of deep learning largely owes to the fast development of computing resources. Most of the deep learning models are trained on high-ended GPUs or CPU clusters. On the other hand, deeper networks impose heavy storage footprint due to the enormous amount of network parameters. For example, the 16-layers VGG involves 528 MBytes of model parameters. Both the high computational and storage cost become impediments to popularize the deep neural networks to scenarios where either memory or computational resources are limited. The great interest to deploy deep learning systems on low-ended devices motives the research in compressing deep models to have smaller computation cost and memory footprints.

Considerable efforts have been mounted to reduce the model size and speed up the inference of deep models. Denil et al.  pointed out that network weights have a significant redundancy, and proposed to reduce the number of parameters by exploiting the linear structure of network \cite{denil}, which motivated a series of low-rank matrix/tensor factorization based compression algorithms, e.g. \cite{svd,svd0,svd1}. Alternatively, multiple studies were devoted to discritizing network weights using vector quantization methods \cite{PQ1,PQ2}, which often outperformed the matrix/tensor factorization based methods \cite{PQ2}.  Han et al. presented the deep compression method that integrates multiple compression methods to achieve a large reduction in model size \cite{DC}. Another line of work for model compression is to restrict network weights to low precision with a few bits.  The advantage of this restriction is that an expensive floating-point multiplication operation can now be replaced by a sequence of cheaper and faster binary bit shift operations. This not only reduces the memory footprints but also accelerates the computation of the network. These approaches work well when pretrained weights are quantized into 4-12 bits~\cite{Lin,dettmers20158,courbariaux2014low,lin2015fixed}. When coming to extremely low bit networks, i.e. only one or two bits are used to represent weights~\cite{BC,hubara2016binarized,cheng2015training}, they only work well on simple datasets (e.g. MNIST and CIFAR10), and usually incur a large loss on challenging datasets like ImageNet.

In this work, we focus on compressing and accelerating deep neural networks with extremely low bits weights, and present a unified strategy for learning such low bits networks. We overcome the limitation of the existing approaches by formulating it as a discretely constrained non-convex optimization problem, which is usually referred to as mixed integer programs (MIP). Given the NP hard nature of MIPs, we proposed a framework for learning extremely low bit neural network using the technique of alternating direction method of multipliers (ADMM) \cite{ADMM_2}.

The main idea behind our method is to decouple the continuous variables from the discrete constraints using an auxiliary variable in the discrete space. This leads to a convenient form of the objective which is amenable to existing nonconvex optimization algorithms. Unlike previous low bit quantization methods \cite{BC,xnor,TWN} that incorporate an ad-hoc modification of the gradients for continuous weights, we simultaneously optimize in both continuous and discrete spaces, and connect the two solutions using an augmented Lagrangian. This is consistent with the previous observation from \cite{MIP_1} that, by decoupling discrete constraints in MIP, one can use the information from the dual problem through ADMM to obtain a better upper bound. As a result of this reformulation, we can divide the problem of low bits quantized neural network into multiple subproblems which are significantly easier to solve. The main contributions of this paper are summarized as follows:

\begin{itemize}
\item We model the low bits neural network as a discretely constrained nonconvex optimization problem, and introduce auxiliary variables to decouple the continuous weights from the discrete constraints. With the use of ADMM, the originally hard problem are decomposed into several subproblems including proximal step, projection step and dual update.

\item We show how the resulting subproblems can be efficiently solved. We utilize extragradient method to accelerate the convergence of proximal step, and propose an iterative quantization algorithm to solve the projection step. The proposed algorithm enjoys a fast convergence in practice.

\item We apply the proposed method to various well-known convolutional neural networks. Extensive experiments on multiple vision tasks including image classification and object detection demonstrate that the proposed method significantly outperforms the state-of-the-art approaches.
\end{itemize}

\section{Related Work}
Due to the high efficiency in both computation and memory footprints, low bits quantization of deep neural networks have received much attention in the literature. In this section, we have a brief review of the representative techniques. We also give a brief introduction to ADMM algorithm and its nonconvex extension.


\subsection {Low bits quantization of neural network}
The research of low bits quantization of neural network can be traced back to 1990s \cite{old1,old2}. Most of the benefits of low bits quantization, such as memory efficiency and multiplication free, had already been explored in these papers. However, the networks are shallow at that age so these approaches do not verify their validity in deep networks and large scale datasets.

In recent years, with the explosion of deep learning in various tasks, low bits quantization techniques have been revisited. Some early works quantize the pretrained weights with 4-12 bits and find such approximations do not decrease predictive performance  \cite{courbariaux2014low,Lin,dettmers20158,lin2015fixed}. More recent works focus on training extremely low bits network from scratch with binary or ternary weights. Among these works, BinaryConnect \cite{BC} is the most representative one. BinaryConnect directly optimizes the loss of the network with weights $W$ replaced by $\text{sign}(W)$. In order to avoid the zero-gradient problem of sign function, the authors approximate it with the ``hard tanh" function in the backward process. This simple idea inspired many following works. BinaryConnect only achieves good results on simple datasets such as MNIST, CIFAR10 and SVHN, but suffers a large degradation on challenging datasets like ImageNet. 

Many efforts have been devoted to improve the performance of BinaryConnect. For example, Binary Weight Network (BWN) \cite{xnor} proposes to improve the performance of BinaryConnect with a better approximation by introducing scale factors for the weights during binarization. Ternary Weight Network (TWN) \cite{TWN} extends the idea of BWN to network with ternary weights and achieves a better performance. Inspired by BinaryConnect, in order to avoid the zero-gradient problem, both BWN and TWN modify the backward process by applying the gradients of the loss at the quantized weights.

Unlike previous works, we mathematically formulated the low bits quantization problem as a discretely constrained problem and present a unified framework based on ADMM to solve it in an efficient way. We simultaneously optimize the problem in both continuous and discrete space, and the two solutions are closely connected in the learning process.

\subsection {ADMM and its nonconvex extension}
Alternating Direction Method of Multipliers (ADMM) \cite{ADMM_2} is an algorithm that is intended to blend the decomposability of dual ascent with the superior convergence properties of the method of multipliers. The algorithm solves problems in the form:
\begin{eqnarray}
&\min& \  f(\mathbf{x}) + g(\mathbf{z}) \nonumber \\
&\text{s.t.}& A\mathbf{x} + B\mathbf{z} = \mathbf{c}
\end{eqnarray}
with variables $\mathbf{x} \in \mathbb{R}^n$ and $\mathbf{z} \in \mathbb{R}^m$, where $A \in \mathbb{R}^{p\times n}$, $B\in \mathbb{R}^{p\times m}$ and $\mathbf{c}\in \mathbb{R}^p$.

The  augmented Lagrangian of Eq.(1) can be formed as:
{\small
\begin{equation}
L_{\rho}(\mathbf{x}, \mathbf{z}, \mathbf{y}) = f(\mathbf{x}) + g(\mathbf{z}) + \mathbf{y}^T(A\mathbf{x} + B\mathbf{z} - \mathbf{c}) + (\rho/2) \|A\mathbf{x} + B\mathbf{z} - \mathbf{c}\|^2_2
\end{equation}}
where $\mathbf{y}$ is the Lagrangian multipliers, and ADMM consists of three step iterations:
\begin{eqnarray}
&\mathbf{x}^{k+1} := &\mathop{\arg\min}_\mathbf{x} L_{\rho} (\mathbf{x}, \mathbf{z}^k, \mathbf{y}^k) \nonumber \\
&\mathbf{z}^{k+1} :=& \mathop{\arg\min}_\mathbf{z} L_{\rho} (\mathbf{x}^{k+1}, \mathbf{z}, \mathbf{y}^k) \nonumber \\
&\mathbf{y}^{k+1} := &\mathbf{y}^k + \rho(A\mathbf{x}^{k+1} + B\mathbf{z}^{k+1} - \mathbf{c}) \nonumber
\end{eqnarray}

Even though ADMM was originally introduced as a tool for convex optimization problems, it turns out to be a powerful heuristic method even for NP-hard nonconvex problems. Recently, this tool has successfully been used as a heuristic to find approximate solutions to nonconvex mixed program problems \cite{MIP_1,MIP_2}, which is very similar to our problem as noted later. 

\section{The Proposed Method}
Let us first define the notion in this paper. Denote $f(W)$ as the loss function of a normal neural network, where
$W = \{W_1, W_2, \cdots, W_L\}$. $W_i$ denotes the weights of the $i$-th layer in
the network, which for example can be a 4-dimension tensor in convolutional layer or a 2-dimension matrix in fully connected layer. For the simplicity of notation, we regard all the entries in $W_i$ as a $d_i$-dimension vector in $\mathbb{R}^{d_i}$, and take $W$ as the concatenation of these vectors so that $W\in \mathbb{R}^d$ with $d = \sum_i d_i.$

In this work, we concentrate on training extremely low bits quantized neural networks. In specific, the weights of the network are restricted to be either zero or powers of two so that the expensive floating-point multiplication operation can be replaced by cheaper and faster bit shift operation. In this section, we aim to mathematically model this problem and efficiently solve it.

\subsection{Objective function}

Intuitively, training a low bits neural network can be modeled as discretely constrained optimization, or in particular, mixed integer programs. For example, the weights in a {ternary} neural network are restricted to be $-1, 0$ or $+1$. Training such network can be mathematically formulated as mixed integer programs:
\begin{equation}
\min \limits_{W} \ \  f(W)  \nonumber \quad \quad \text{s.t.} \ \   W \in \mathcal{C} = \{-1, 0, +1\}^d \nonumber
\end{equation}

Since the weights are restricted to be zero or powers of two, we have constraints of this form
$$\mathcal{C} = \{-2^N,\cdots, -2^1, -2^0, 0, +2^0, +2^1, \cdots, +2^N\}$$
where $N$ is an integer which determines the number of bits. As in \cite{xnor}, we further introduce a scaling factor $\alpha$ to the constraints, i.e., instead of requiring $\mathcal{C} = \{\cdots, -2, -1, 0, +1, +2. \cdots\}$, we simply restrict $\mathcal{C}$ to $\mathcal{C} = \{\cdots, -2\alpha, -\alpha, 0, +\alpha, +2\alpha, \cdots\}$ with an arbitrary scaling factor $\alpha > 0$ that is strictly positive. It is worthy noting that the scale factor $\alpha$ in various layers can be different. In other words, for a neural network with $L$ layers, we actually introduce $L$ different scaling factors $\{\alpha_1, \alpha_2, \cdots, \alpha_L\}$. Formally, the objective function of low bits quantized neural networks can be formulated as: 
\begin{eqnarray} \label{eq3}
&\min \limits_{W}& f(W) \nonumber \\ 
&\text{s.t.}& W\in \mathcal{C} = \mathcal{C}_1 \times \mathcal{C}_2 \times \cdots \times \mathcal{C}_L
\end{eqnarray}
where $\mathcal{C}_i=\{0,\pm\alpha_i, \pm2\alpha_i, \cdots, \pm2^N\alpha_i \}$ and $\alpha_i > 0$. We emphasize that the scaling factor $\alpha_i$ in each layer doesn't incur more computation to the convolutional operator, because it can be multiplied after the efficient convolution with $\{0,\pm1, \pm2, \cdots, \pm2^N \}$ done. 

\begin{figure}[t]

\includegraphics[scale=0.7]{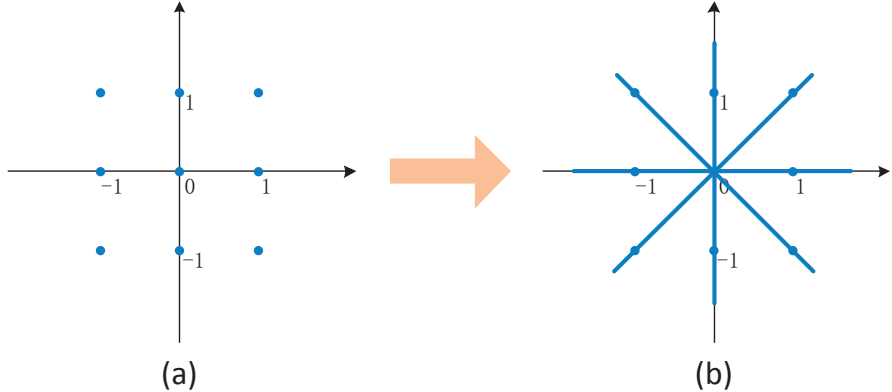}

\caption{In ternary neural network, scaling factor expands the constrained space from (a) nice discrete points to (b) four lines in the space (two dimensional space as an example).}\label{fig:scale}
\end{figure}

From the perspective of constrained optimization, the scaling factor $\alpha_i$ helps to expand the constraint space.  As an example, Fig.1 gives an illustration of how it works for ternary network. In two dimensional space, for constraint ${\{-1, 0, +1\}}$, the possible solutions of ternary neural network are nine discrete points in the space. In contrast, with the scaling factor added, the constrained space is expanded to be four lines in the space. This large expansion of the constrained space will make the optimization easier. 

\subsection{Decouple with ADMM}
The optimization in Eq.(3) is NP-hard in general because the weights are constrained in a discrete space. Most previous works try to directly train low bits models to minimize the loss. For example, BinaryConnect \cite{BC} replace the weights $W$ with $\text{sign}(W)$ in the forward process so that the constraints will be automatically satisfied. Since the gradients of $\text{sign}(W)$ to $W$ is zero everywhere, the authors replace the sign function with ``hard tanh" in the backward process. The same idea is also adopted by BWN \cite{xnor} and TWN \cite{TWN}. However, as indicated in \cite{cai2017deep}, the use of different forward and backward approximations causes the mismatch of gradient, which makes the optimization instable.

We overcome the limitation of previous approaches by converting the problem into a form which is suitable to existing nonconvex optimization techniques.  We introduce an auxiliary variable which is subject to the discrete restriction and equal to original variable. This is used with ADMM, which will result in the effect that the discrete variables being decoupled when we consider their minimization. Our basic idea is largely inspired by recent successful application of ADMM in mixed integer programs \cite{MIP_1,MIP_2}. 

First of all, defining an indicator function $I_{\mathcal{C}}$ for whether $W\in \mathcal{C}$, the objective in Eq.\eqref{eq3} can be written as 
\begin{equation} \label{eq4}
{\min \limits_{W}} \quad f(W) + I_{\mathcal{C}}(W)
\end{equation}
where $I_{\mathcal{C}}(W)=0$ if $W\in \mathcal{C}$, otherwise $I_{\mathcal{C}}(W)=+\infty$. 

By introducing an auxiliary variable $G$, we can rewrite the optimization in Eq.\eqref{eq4} with an extra equality constraint so that the weights is constrained to be equal to the discrete variable, but not subject to that restriction. In detail, the objective can be reformulated as:
\begin{eqnarray} \label{eq5}
&{\min \limits_{W, G}}& \  f(W) + I_{\mathcal{C}}(G) \nonumber \\
&\text{s.t.}&\  W = G
\end{eqnarray}

Now we are considering a nonconvex optimization with convex linear constraints. Problems of such form can be conveniently solved with ADMM. The augmented Lagrange of Eq.\eqref{eq5}, for parameter $\rho > 0$, can be formulated as: 
\begin{equation} \label{eq6}
L_{\rho}(W,G,\mu) =  f(W) + I_{\mathcal{C}}(G) + \frac{\rho}{2}\|W-G\|^2 + \left<\mu,W - G\right>
\end{equation}
where $\mu$ denotes the Lagrangian multipliers and  $\left<\cdot, \cdot\right>$ denotes the inner product of two vectors. With some basic collection of terms and a change of variable $\lambda=(1/\rho)\mu$, Eq.\eqref{eq6} can be equivalently formed as:
\begin{equation} \label{eq7}
L_{\rho}(W,G,\lambda) =  f(W) + I_{\mathcal{C}}(G) + \frac{\rho}{2}\|W-G+\lambda\|^2 - \frac{\rho}{2}\|\lambda\|^2
\end{equation}

Following the standard process of ADMM,  this problem can be solved by repeating the following iterations:
\begin{eqnarray}
&W^{k+1} :=& \mathop{\arg\min}_W \ L_{\rho}(W, G^k, \lambda^k)\\
&G^{k+1} :=& \mathop{\arg\min}_G \ L_{\rho}(W^{k+1}, G, \lambda^k)\\
&\lambda^{k+1} :=& \lambda^{k} + W^{k+1} - G^{k+1}
\end{eqnarray}
which is respectively the proximal step, projection step and dual update. 

Unlike previous works, we simultaneously optimize the problem in both continuous space (i.e., proximal step) and discrete space (i.e., projection step), and the two solutions are brought together by ADMM in the learning process.

\subsection{Algorithm subroutines}
In this section, we elaborate on how the consequent subproblems in the above algorithm can be efficiently solved. 
\subsubsection{Proximal step}
For the proximal step, we optimize in the continuous space. Formally, we need to find the weights that minimize
\begin{equation}
L_{\rho}(W, G^k, \lambda^k) = f(W) + \frac{\rho}{2} \|W-G^k + \lambda^k\|^2
\end{equation}

Due to the decouple of ADMM, we are dealing with an unconstrained objective here. The loss can be interpreted as a normal neural network with a special regularization. Naturally, this problem can be solved with standard gradient decent method. It is easy to obtain the gradient with respect to the weights $W$:
$$\partial_W L = \partial_W f + \rho(W-G^k + \lambda^k)$$

However, we find the vanilla gradient descent method converges slowly in this problem. Since the second quadratic term occupies a large proportion of the whole lost, SGD will quickly pull the optimizer to the currently quantized weights so that the second term vanishes, and stack in that point. This results in a suboptimal solution since the loss of neural network is not sufficiently optimized. 


To overcome this challenge, we resort to the extragradient method \cite{extra}. An iteration of the extragradient method consists of two very simple steps, prediction and correction:
\begin{align*}
  W^{(p)} & := W - \beta_p \partial_W L(W), \\
  W^{(c)} &:= W - \beta_c \partial_W L(W^{(p)})
\end{align*}
where $\beta_p$ and $\beta_c$ are the learning rates. A distinguished feature of the extragradient method is the use of an additional gradient step which can be seen as a guide during the optimization process. Particularly, this additional iteration allows to foresee the geometry of the problem and take the curvature information into account, which leads to a better convergency than standard gradient descent \cite{extra2}. Specific to our problem, there is a more intuitive understanding of the above iterations. For the prediction step, the algorithm will quickly move to a point close to $G^k - \lambda^k$ so that the loss of quadratic regularization vanishes. Then in the correction step, the algorithm moves another step which tries to minimize the loss of neural network $f(W)$. These two steps avoid the algorithm stacking into a less valuable local minima. In practice, we find this extragradient method largely accelerate the convergence of the algorithm.

A key observation of \eqref{eq12} is that while minimizing over $G$, all the components $G_i$ are decoupled, therefore the auxiliary variables of each layer can be optimized independently. Recall that $W_i, G_i, \lambda_i, \mathcal{C}_i$ denote the weights, auxiliary variables, Lagrangian multipliers and constraints of the $i$-th layer respectively.  We are essentially looking for the Euclidean projection of $(W^{k+1}_i + \lambda^k_i)$ onto a discrete set $\mathcal{C}_i$. Since the constraint is discrete and nonconvex, this optimization is nontrivial. 

For convenience, we denote $(W^{k+1}_i + {\lambda^k_i})$ as $V_i$. The projection of $V_i$ onto $\mathcal{C}_i$ can be formulated as
\begin{eqnarray}
&\min \limits_{G_i, \alpha_i}& \|V_i - G_i\|^2 \nonumber \\
&\text{s.t.} & G_i \in \{0,\pm\alpha_i, \pm2\alpha_i, \cdots, \pm2^N\alpha_i \}^{d_i}
\end{eqnarray}

Taking the scaling factor away from the constraints, the objective can be equivalently formulated as:
\begin{eqnarray}
&\min \limits_{Q_i, \alpha_i}& \|V_i - \alpha_i\cdot Q_i\|^2 \nonumber \\
&\text{s.t.} & Q_i \in \{0,\pm1, \pm2, \cdots, \pm2^N\}^{d_i}
\end{eqnarray}

We propose an iterative quantization method to solve this problem. The algorithm alternates between optimizing $\alpha_i$ with $Q_i$ fixed and optimizing $Q_i$ with $\alpha_i$ fixed. In specific, with $Q_i$ fixed, the problem becomes an univariate optimization. The optimal $\alpha_i$ can be easily obtained as
\begin{equation}
\alpha_i = \frac{V_i^TQ_i}{Q_i^TQ_i}
\end{equation}

With $\alpha_i$ fixed, the optimal $Q_i$ is actually the projection of $\frac{V_i}{\alpha_i}$ onto $\{0,\pm1, \pm2, \cdots, \pm2^N\}$, namely, 
\begin{equation}
Q_i = \Pi_{\{0,\pm1, \pm2, \cdots, \pm2^N\}}\left(\frac{V_i}{\alpha_i}\right)
\end{equation}
where $\Pi$ denotes the projection operator. Moreover, the projection onto a discrete set is simply the closest point in it. 

This iterative quantization algorithm is guaranteed to converge to a local minimum since we can get a decrease of loss in each step. In practice, we also find such a simple algorithm converges very fast. In most cases, we only need less than five iterations to get a stable solution.

\subsubsection{Dual update}
In ADMM, dual update is actually gradient ascent in the dual space \cite{ADMM_2}. The iterate $\lambda^{k+1}$ in Eq.(10) can be interpreted as a scaled dual variable, or as the running sum of the error values $W^{k+1}-G^{k+1}$.

\section{Experiments}
In order to verify the effectiveness of the proposed algorithm, in this section we evaluate it on two benchmarks: ImageNet for image classification and Pascal VOC for object detection. 

\subsection{Image Classification}
To evaluate the performance of our proposed method on image recognition task, we perform extensive experiments on the large scale benchmark ImageNet (ILSVRC2012), which is one of the most challenging image classification benchmarks. ImageNet dataset has about 1.2 million training images and 50 thousand validation images, and these images cover 1000 object classes. We comprehensively evaluate our method on almost all well-known deep CNN architectures, including AlexNet \cite{AlexNet}, VGG-16 \cite{VGG}, ResNet-18 \cite{ResNet}, ResNet-50 \cite{ResNet} and GoogleNet \cite{GoogleNet}.

\subsubsection{Experimental setup}
In the ImageNet experiments, all the images are resized to $256\times 256$. The images are then randomly clipped to $224\times224$ patches with mean subtraction and randomly flipping. No other data augmentation tricks are used in the learning process. We report both the top-1 and top-5 classification accurate rates on the validation set, using single-view testing (single-crop on central patch only).

We study different kinds of bit width for weight quantization. Specifically, we tried binary quantization, ternary quantization, one-bit shift quantization and two-bits shift quantization. For one-bit shift quantization, the weights are restricted to be \{-2$a$, -$a$, 0, +$a$, +2$a$\}, which we denote as \{-2, +2\} in the comparison. Similarly, two-bits shift quantization are denoted as \{-4, +4\}. Binary quantization and ternary quantization need one bit and two bits to represent one weight respectively. Both \{-2, +2\} quantization and  \{-4, +4\} quantization need three bits to represent one weight. 

For binary and ternary quantization, we compare the proposed algorithm with the state-of-the-art approaches Binary Weight Network (BWN) \cite{xnor} and Ternary Weight Network (TWN) \cite{TWN}. Both BWN\footnote{\url{https://github.com/allenai/XNOR-Net}} and TWN\footnote{\url{https://github.com/fengfu-chris/caffe-twns}} release their source code so we can evaluate their performance on different network architectures. Our method is implemented with Caffe \cite{caffe}. The referenced full precision CNN models VGG-16, ResNet-50 and GoogleNet are taken from the Caffe model zoo\footnote{\url{https://github.com/BVLC/caffe/wiki/Model-Zoo}}. 
\begin{table*}[t]
\centering
\begin{tabular}{c|c||c|c|c|c|c|c|c}
\hline 
&Accuracy & Binary & BWN & Ternary & TWN & \{-2, +2\} &  \{-4, +4\} & Full Precision \\
\hline
\hline
\multirow{2}{*} {AlexNet} & Top-1 & \textbf{0.570} & 0.568 & \textbf{0.582} & 0.575 & 0.592 & 0.600 & 0.600 \\
\cline{2-9}
& Top-5 & \textbf{0.797} & 0.794 & \textbf{0.806} & 0.798 & 0.818 & 0.822 & 0.824 \\
\hline
\hline
\multirow{2}{*} {VGG-16} & Top-1 & \textbf{0.689} & 0.678 & \textbf{0.700} & 0.691 & 0.717 & 0.722 &  0.711 \\
\cline{2-9}
& Top-5 & \textbf{0.887} & 0.881 & \textbf{0.896} & 0.890 & 0.907 & 0.909 & 0.899 \\
\hline
\end{tabular}
\caption{Accuracy of AlexNet and VGG-16 on ImageNet classification.}
\end{table*}

\begin{table*}[t]
\centering
\begin{tabular}{c|c||c|c|c|c|c|c|c}
\hline 
&Accuracy & Binary & BWN & Ternary & TWN & \{-2, +2\} &  \{-4, +4\} & Full Precision \\
\hline
\hline
\multirow{2}{*} {Resnet-18} & Top-1 & \textbf{0.648} & 0.608 & \textbf{0.670} & 0.618 & 0.675 & 0.680 & 0.691 \\
\cline{2-9}
& Top-5 & \textbf{0.862} & 0.830 & \textbf{0.875} & 0.842 & 0.879 & 0.883 & 0.890 \\
\hline
\hline
\multirow{2}{*} {Resnet-50} & Top-1 & \textbf{0.687} & 0.639 & \textbf{0.725} & 0.656 & 0.739 & 0.740 & 0.753 \\
\cline{2-9}
& Top-5 & \textbf{0.886} & 0.851 & \textbf{0.907} & 0.865 & 0.915 & 0.916 & 0.922 \\
\hline
\end{tabular}
\caption{Accuracy of ResNet-18 and ResNet-50 on ImageNet classification.}
\end{table*}

\begin{table*}[t]
\centering
\begin{tabular}{c||c|c|c|c|c|c|c}
\hline 
Accuracy & Binary & BWN & Ternary & TWN & \{-2, +2\} &  \{-4, +4\} & Full Precision \\
\hline
\hline
Top-1 & \textbf{0.603} & 0.590 & \textbf{0.631} & 0.612 & 0.659 & 0.663 & 0.687 \\
\hline
Top-5 & \textbf{0.832} & 0.824 & \textbf{0.854} & 0.841 & 0.873 & 0.875 & 0.889 \\
\hline
\end{tabular}
\caption{Accuracy of GoogleNet on ImageNet classification.}
\end{table*}


\subsubsection{Results on AlexNet and VGG-16}
AlexNet and VGG-16 are ``old fashion" CNN architectures. AlexNet consists of 5 convolutional layers and 3 fully-connected layers. VGG-16 uses much wider and deeper structure than AlexNet, with 13 convolutional layers and 3 fully-connected layers. Table 1 demonstrates the comparison results on these two networks. For fair comparison with BWN, we report the performance of the batch normalization \cite{BN} version of AlexNet. The accuracy of the improved AlexNet is higher than the original one (Top-1 60.0\% vs. 57.4\%, Top-5 82.4\% vs. 80.4\%).

On these two architectures, the proposed algorithm achieves a lossless compression with only 3 bits compared with the full precision references. For \{-2, +2\} and \{-4, +4\} quantization, the performance of the our quantized networks is even better than the original full precision network on VGG-16. Similar results are observed in BinaryConnect on small datasets. This is because discrete weights could provide a form of regularization which can help to generalize better. These results also imply the heavy redundancy of the parameters in full precision AlexNet and VGG-16 models. This finding is consistent with that in other studies such as SqueezeNet \cite{Squeezenet}. In SqueezeNet, the authors suggest that one can achieve AlexNet-level accuracy on ImageNet with 50x fewer parameters.

Our binary quantization and ternary quantization slightly outperforms BWN and TWN on these two architectures. Comparing the accuracy of ternary quantization and binary quantization, we find that ternary network consistently works better than binary network. We also emphasize that the ternary network is more computing efficient than binary network because of the existence of many zero entries in the weights, as indicated in \cite{venkatesh2017accelerating}.  

\subsubsection{Results on ResNet}
The results on ResNet-18 are shown in Table 2. ResNet-18 has 18 convolutional layers with shortcut connections. For the proposed method, both the binary and ternary quantization substantially outperform their competitors on this architecture. For example, our binary network outperforms BWN by 4 points in top-1 accuracy and 3.2 points in top-5 accuracy. The proposed ternary quantization outperforms TWN by 5.2 points and 3.3 points in top-1 and top-5 accuracy respectively. All these gaps are significant on ImageNet. We also observe over two percent improvement for our ternary quantization over binary quantization.

The effectiveness of our method is also verified on very deep convolutional network such as ResNet-50. Besides significantly increased network depth, ResNet-50 has a more complex network architecture than ResNet-18. Table 2 details the results on ResNet-50. It is easy to observe the similar trends as in ResNet-18. Our method is considerably better than the compared BWN and TWN. For example, our binary quantization obtains about 5 points improvement on top-1 accuracy over BWN.

For both ResNet-18 and ResNet-50, there is a more noticeable gap between the low bits quantized networks and full precision reference. Different from AlexNet and VGG-16, on ResNet we notice about 1 point gap in top-1 accuracy between \{-4, +4\} quantized network and full precision reference. These results suggest that training extremely low bits quantized network is easier for AlexNet and VGG than for ResNet, which also implies the parameters in AlexNet and VGG-16 are more redundant than those in ResNet-18 and ResNet-50.  

\subsubsection{Results on GoogleNet} The results on GoogleNet are illustrated in Table 3. GoogleNet is a 22 layers deep network, organized in the form of the ``Inception module". Similar to ResNet, GoogleNet is more compact than AlexNet and VGG-16, so it will be more difficult to compress it. There exists a gap of more than 2 points in top-1 accuracy between \{-4, +4\} quantized network and full precision version. The loss of binary quantization is more significant, which reaches 8 points in top-1 accuracy. Despite this, our method stills outperforms BWN\footnote{Note that the GoogleNet used in BWN paper is an improved variant of the original version used in this paper.} and TWN on this network.

\subsubsection{Compare with the most recent works}
To our knowledge, Trained Ternary Quantization (TTN) \cite{TTN} and Incremental Network Quantization (INQ) \cite{INQ} are two of the most recent published works on low bits quantization of deep neural network. Instead of quantizing the ternary weights to be $\{-\alpha, 0, +\alpha\}$, TTN makes it less restrictive as $\{-\alpha, 0, +\beta\}$. Note that our method can be easily extended to deal with constraints of such form. Nevertheless, the computation of such form of ternary network is less efficient than the original one. As an example, for fast implementation the inner product  between vector $\textbf{x}$ and vector $(-\alpha, -\alpha, 0, +\beta)$ will be decomposed as $\beta \textbf{x} \cdot (0,0,0,1)-\alpha\textbf{x} \cdot (1,1,0,0)$, having to do two floating-point multiplications with $\alpha$ and $\beta$. 

Since TTN only reports its results on AlexNet and ResNet-18, we compare the performance on these two architectures. Detailed results are summarized in Table 4 and Table 5. Our approach performs better than TTN on AlexNet (the results of ternary INQ on AlexNet is not available), and better than both TTN and INQ on ResNet-18. INQ shows more results on 5-bits networks in the paper. For example, the reported top-1 and top-5 accuracy of ResNet-50 with 5-bits are 73.2\% and 91.2\% \cite{INQ}. In contrast, our method achieves such accuracy with only 3 bits.

\begin{table}[ht]
\centering
\begin{tabular}{c||c|c}
\hline 
Method & Top-1 accuracy  & Top-5 accuracy  \\
\hline
\hline
TTN \cite{TTN} & 0.575 & 0.797 \\
\hline
Ours (Ternary) & \textbf{0.582} & \textbf{0.806} \\
\hline
\end{tabular}
\caption{Comparison with TTN on AlexNet.}
\end{table}

\begin{table}[ht]
\centering
\begin{tabular}{c||c|c}
\hline 
Method & Top-1 accuracy  & Top-5 accuracy  \\
\hline
\hline
TTN \cite{TTN} & 0.666 & 0.872 \\
\hline
INQ \cite{INQ} & 0.660 & 0.871 \\
\hline
Ours (Ternary) & \textbf{0.670} & \textbf{0.875} \\
\hline
\end{tabular}
\caption{Comparison with TTN and INQ on  ResNet-18.}
\end{table}

\subsubsection{INT8 quantized 1$\times$1 kernel}
We notice the extremely low bits quantization of GoogleNet suffers a large degradation. We guess this may be due to the 1$\times$1 kernel in each inception.  In order to verify this point, we perform another experiment on GoogleNet. In this version, the 1$\times$1 kernels in the network are quantized with relatively more bits, i.e., INT8, and kernels of other size are quantized as usual. Table 6 shows the results.

By comparing the results in Table 6 and those in Table 3, we observe a considerable improvement, especially for binary and ternary quantization. As we have discussed, discrete weights can be interpret as a strong regularizer to the network. However, the parameters in 1$\times$1 kernel is much less than those in other kernels. Imposing a very strong regularizer to such kernels may lead to underfitting of the network. These results suggest that we should quantize different parts of the networks with different bit width in practice. Letting the algorithm automatically determine the bit width will be our future work.


\begin{table}[t]
\centering
\begin{tabular}{c||c|c|c|c|c}
\hline 
Acc. & Binary & Ternary & \{-2, +2\} &  \{-4, +4\} & Full \\
\hline
\hline
Top-1 & {0.654} & {0.667} & 0.674 & 0.676 & 0.687 \\
\hline
Top-5 & {0.867} & {0.877} & 0.883 & 0.883 & 0.889 \\
\hline
\end{tabular}
\caption{Accuracy of GoogleNet. 1$\times$1 kernels are quantized with INT8.}
\end{table}

\subsection{Object Detection}
In order to evaluate the proposed method on object detection task, we apply it to the state of arts detection framework SSD \cite{SSD}. The models are trained on Pascal VOC2007 and VOC2012 train dataset, and tested on Pascal VOC 2007 test dataset. For SSD, we adopt the open implementation released by the authors\footnote{\url{https://github.com/weiliu89/caffe/tree/ssd}}. In all experiments, we follow the same setting as in \cite{SSD} and the input images are resized to $300\times300$. 

The proposed method are evaluated on two base models, i.e., VGG-16 and Darknet reference model. Both base networks are pre-trained on ImageNet dataset. The VGG-16 network here is a variant of original one \cite{VGG}. In detail, the fc6 and fc7 are converted to convolutional layers with $1\times 1$ kernel, and the fc8 layer is removed. The parameters of fc6 and fc7 are also subsampled. The darknet reference model is borrowed from YOLO \cite{yolo}, which is another fast detection framework. Darknet is designed to be small yet power, which attains comparable accuracy performance as AlexNet but only with about 10$\%$ of the parameters.  We utilize the base darknet model downloaded from the website\footnote{\url{https://pjreddie.com/darknet/imagenet/}}. 

To the best of our knowledge, there is no other works on low bits quantization applied their algorithms to the object detection tasks. We compare our quantized network with full precision network in this experiment. We only implement ternary and \{-4,+4\} quantization for this experiment. Darknet has utilized many 1$\times$1 kernels as in GoogleNet to accelerate the inference process.  We implement two versions of Darknet. In the first version, the 1$\times$1 kernels are also quantized as usual, while in the second version these kernels are quantized with INT8. Table 7 shows the mean average precision (mAP) on both models. 

For \{-4,+4\} quantization, we find that the mAP of both modes are very close to the full precision version. On VGG16+SSD, we only suffer a loss of 0.002 in mAP.  Comparing two versions of Darknet+SSD, the first version achieves a mAP of 0.624, and the second version obtains a improvement of 1.5 points.  For ternary quantization, the accuracy degradation of Darknet+SSD is larger than VGG16+SSD, because the parameters of Darknet is less redundant than VGG-16. All these results indicate that our proposed method is also effective on the object detection tasks. 

\begin{table}[t]
\centering
\begin{tabular}{c||c|c}
\hline 
mAP  &  Darknet+SSD &  VGG16+SSD  \\
\hline
\hline
Ternary & 0.609 (0.621) & 0.762  \\
\hline
\{-4, +4\} & 0.624 (0.639) & 0.776  \\
\hline
Full Precision & 0.642 & 0.778 \\
\hline
\end{tabular}
\caption{mAP of VGG16+SSD and Darknet+SSD on Pascal VOC 2007}
\end{table}

\section{Conclusion}
This work focused on compression and acceleration of deep neural networks with extremely low bits weight. Inspired by the efficient heuristics proposed to solve mixed integer programs, we proposed to learn low bits quantized neural network in the framework of  ADMM. We decoupled the continuous parameters from the discrete constraints of network, and cast the original hard problem into several subproblems. We proposed to solve these subproblems using extragradient and iterative quantization algorithms that lead to considerably faster convergency compared to conventional optimization methods. Extensive experiments on convolutional neural network for image recognition and object detection have shown the effectiveness of the proposed method.

{\footnotesize
\bibliographystyle{IEEEtran}
\bibliography{bibfile1}
}

\end{document}